\documentclass[conference]{IEEEtran}
\usepackage{cite}
\usepackage{amsmath,amssymb,amsfonts}
\usepackage{algorithmic}
\usepackage{graphicx}
\usepackage{textcomp}
\usepackage{xcolor}
\usepackage{tikz}
\usepackage{latexsym}
\usepackage{multirow}
\usepackage[utf8]{inputenc}
\usepackage{booktabs}
\usepackage{hyperref}
\usetikzlibrary{decorations.pathmorphing}
\usepackage[acronym, nonumberlist]{glossaries}

\def\BibTeX{{\rm B\kern-.05em{\sc i\kern-.025em b}\kern-.08em
    T\kern-.1667em\lower.7ex\hbox{E}\kern-.125emX}}
\begin{document}

\title{Towards meta-learning for multi-target \\regression problems}

\author{
    \IEEEauthorblockN{
    Gabriel J. Aguiar \IEEEauthorrefmark{1},
    Everton J. Santana \IEEEauthorrefmark{1},
    Saulo M. Mastelini \IEEEauthorrefmark{2},
    Rafael G. Mantovani \IEEEauthorrefmark{3},
    Sylvio Barbon Jr \IEEEauthorrefmark{1}}

    \IEEEauthorblockA{\IEEEauthorrefmark{1} Computer Science Department, Londrina State University, Londrina - PR, Brazil} 

    \IEEEauthorblockA{\IEEEauthorrefmark{2} Institute of Mathematics and Computer Sciences, University of S\~ao Paulo (ICMC/USP), S\~ao Carlos - SP, Brazil} 

    \IEEEauthorblockA{\IEEEauthorrefmark{3} Computer Engineering Department, Federal Technology University, Campus of Apucarana - PR, Brazil
    \\ E-mail: gjonas@uel.br, evertonsantana@uel.br, mastelini@usp.br, rafaelmantovani@utfpr.edu.br, barbon@uel.br}
}

\maketitle

\begin{abstract}
Several multi-target regression 
methods were developed in the last years aiming at improving predictive performance by exploring inter-target correlation within the problem. However, none of these methods outperforms the others for all problems. This motivates the development of automatic approaches to recommend the most suitable multi-target regression method. 
In this paper, we propose a meta-learning system to recommend the best predictive 
method for a given multi-target regression problem. We performed experiments with a meta-dataset generated by a total of 648 synthetic datasets. These datasets were created to explore distinct inter-targets characteristics toward recommending the most promising 
method. In experiments, we evaluated four different 
algorithms with different biases as meta-learners.
Our meta-dataset is composed of 58 meta-features, based on: statistical information, correlation characteristics, linear landmarking, from the distribution and smoothness of the data, and has four different meta-labels.
Results showed that induced meta-models were able to recommend the best 
method for different base level datasets with a balanced accuracy superior to $70$\% using a Random Forest meta-model, which statistically outperformed the meta-learning baselines. 

\end{abstract}


\newacronym{ml}{ML}{Machine Learning}
\newacronym{mtl}{MtL}{Meta-learning}
\newacronym{oob}{OOB}{Out-of-Bag}

\newacronym{ann}{ANN}{Artificial Neural Network}
\newacronym{svm}{SVM}{Support Vector Machine}
\newacronym{rf}{RF}{Random Forest}
\newacronym{xgb}{XGBoost}{Extreme Gradient Boosting}
\newacronym{nb}{NB}{Naive Bayes}
\newacronym{cd}{CD}{Critical Difference}
\newacronym{cv}{CV}{Cross-Validation}

\newacronym{motc}{MOTC}{Multi-output Tree Chaining}
\newacronym{st}{ST}{Single-target}
\newacronym{sst}{SST}{Stacking Single Target}
\newacronym{erc}{ERC}{Ensemble of Regressor Chains}
\newacronym{mtr}{MTR}{Multi-target Regression}
\newacronym{mlc}{MLC}{Multi-label Classification}

\newacronym{rrmse}{aRRMSE}{Average Relative Root Mean Square Error}
\newacronym{rmse}{RMSE}{Root Mean Square Error}
\newacronym{cor}{COR}{Correlation}
\newacronym{lin}{LIN}{Linearity}

\newcommand{\todo}[1]{\textcolor{red}{\begin{flushleft} $\blacktriangleleft$\texttt{#1}$\blacktriangleright$ \end{flushleft}}}


\section{Introduction}

\acrfull{ml} approaches have been providing significant advances in understanding and modeling problems from the broadest knowledge fields. A considerable part of the \acrshort{ml} solutions takes advantage of supervised learning algorithms which explore the information, i.e., input and prediction target, from the problem data to learn a pattern. However, data from several real problems present more than one target. In this case, when a dataset presents multiple continuous targets, we call it a \textit{multi-target regression problem}. 

Currently, there are several methods in the literature addressing this type of problem. The most straightforward approach, referred to as \acrfull{st} regression, is to create a single model for each target disregarding the possible inter-target correlation. \acrfull{mtr} is an alternative approach that, besides using the original input features, exploits the statistical correlation among the outputs. The \acrshort{mtr} methods have been applied to solve many problems~\cite{Tsoumakas2014, Levatic2014, sanatana2018,Mastelini2018,santana2019stock}, leading to improvement in the predictive performance over \acrshort{st} methods. However, each method has specific characteristics and has been effective for different problems. 

Selecting the most suitable algorithm for a given problem requires extensive experimental evaluation, which demands massive computational resources (particularly processing time) and specialists \cite{bilalli2017predictive,cunha2018metalearning}. On the other hand, a \acrshort{mtr} method could be automatically selected when addressed as an output in an algorithm selection (or recommendation) problem by ~\acrfull{mtl}~\cite{Brazdil:2009}.

The~\acrshort{mtl} core concept is to use the knowledge acquired from previous similar problems to recommend the most suitable algorithm, for a new unseen dataset. In the last years, \acrshort{mtl} has been employed in different contexts, such as tasks to select \cite{ali2006meta}, rank \cite{rossi2014metastream} and predict \cite{reif2012meta} the performance of \acrshort{ml} algorithms and employing them on a new dataset. 

Our hypothesis holds that \acrshort{mtl} can be applied to \acrshort{mtr} problems and recommend the most suitable method for new unseen problems. Thus, in this study, we propose a recommendation system able to predict the best \acrshort{mtr} method for a new dataset. For such, experiments were carried out with meta-datasets generated with a total of $648$ synthetic regression problems, also generated to explore the different inter-targets characteristics. In the experiments, the \acrshort{st} approach and three \acrshort{mtr} methods were evaluated:~\acrfull{sst}~\cite{Spyromitros2016},~\acrfull{motc}~\cite{Mastelini2018} and ~\acrfull{erc}~\cite{Spyromitros2016}. Thus, the meta-knowledge was generated with different datasets, with different biases, often used for multi-target benchmarking~\cite{mastelini2018benchmarking}. In the experiments, we compared \acrfull{nb}, \acrfull{rf}, \acrfull{xgb} and \acrfull{svm} as meta-learners using their default hyperparameter values.

This paper is structured as follows. Section~\ref{sec:related} presents the background on using \acrshort{mtl}~for~\acrshort{mtr}; section \ref{sec:materials} describes the experimental methodology; the results are discussed in section~\ref{sec:results}; finally, the conclusions and future work are presented.


\section{Background}
\label{sec:related}

Many \acrshort{ml} algorithms have been proposed for different prediction tasks. However, the '\textit{No free lunch}' theorem~\cite{wolpert1996lack} states there is no one algorithm suitable for every dataset. A possible solution is to recommend the best algorithm for each problem.

The notion of algorithm recommendation problems was introduced in~\cite{Rice:1976}, grounded on selecting one algorithm from a portfolio of options. Given a set of datasets $\mathcal{P}$ composed of instances from a distribution $\mathcal{Q}$; a set of algorithms $\mathcal{A}$; and a performance measure  $\mathcal{M}$\::\:$\mathcal{P}$\:$\times$\:$\mathcal{A}\rightarrow\mathbb{R}$; 
the algorithm recommendation problem is to find a mapping $m$\::\:$\mathcal{P}\rightarrow\mathcal{A}$ that optimizes the expected performance measure for the instance problems described in $\mathcal{Q}$.

In practice, there are some alternatives to induce this mapping between algorithms 
and datasets/problems: 
 one of them is through the \acrfull{mtl}~\cite{Brazdil:2009}.
The core concept of ~\acrshort{mtl} is to exploit past learning experiences in a particular type of task and solutions by adapting learning algorithms and data mining processes.
This is done by extracting features from a dataset, named as meta-features, to represent a dataset and the performance of the \acrshort{ml} algorithms when applied on it. The relation between \textit{meta-features} and the \acrshort{ml} performance provides information to select the most suitable algorithm for new datasets. Thus, \acrshort{ml} algorithms are applied to a meta-dataset, whose examples are described in terms of meta-features, to induce a meta-model.

In the last years, \acrshort{mtl} has been used for: 
algorithm selection~\cite{ferrari2015clustering}, 
segmentation algorithm recommendation~\cite{campos2016meta}, and hyperparameter tuning~\cite{Mantovani:2019}.


\subsection{Multi-target regression}

\acrfull{mtr} is related to the problems with multiple continuous outputs. In this way, to solve these problems a function or a collection of functions $\mathcal{H}$ that models the relationship from input ($\mathcal{X}$) to output ($\mathcal{Y}$) is created. If $\mathcal{X}$ is composed of $m$ input variables and $\mathcal{Y}$ has $d$ targets, the prediction problem can be stated as:

\begin{equation}
    \mathcal{H}: {\mathcal{X}}_{1 \dots m} \xrightarrow\:{\mathcal{Y}}_{1 \dots d}
\end{equation}

Then, for each vector that belongs to $\mathcal{X}$, $\mathcal{H}$ is capable of predicting an output vector that is the best approximation of the true output vector \cite{Spyromitros2016}.

\acrshort{mtr} methods might use one of two main procedures: Algorithm Adaptation or Problem Transformation \cite{Borchani2015}. The first one adapts well-known algorithms, such as: \acrfullpl{ann}; \acrfull{rf} and \acrfullpl{svm}, to deal with multiple outputs, modeling the problem at once. On the other hand, problem transformation methods modify the original input task aiming at exploring the correlation among the targets. 
Spyromitros-Xioufis \textit{et al.}~\cite{Spyromitros2016} proposed two problem transformation methods that contributed notably to the area: ~\acrfull{sst} and \acrfull{erc}. The \acrshort{sst} method builds one model for each target $d$, which are iteratively stacked to the input, and induced new $d$ models over the augmented input. The prediction of these last models are the final predictions.

Differently, the~\acrshort{erc} method creates regressors based on a different order of the targets. For each order, models are trained sequentially: the model that is trained for the second response considers the model trained for the first one. Both models are used in the induction of the third regressor, and so forth. In the end, for each target, the prediction is the average of the predictions of the trained regressors.

These both methods inspired the development of new \acrshort{mtr} methods~\cite{melki2017multi, Mastelini2017, Moyano2017}. One of them, the \acrfull{motc}~\cite{Mastelini2018}, is a method that requires less memory and training time than ERC, besides generating an interpretation of the targets' dependencies. It creates regressors from a tree built based on correlation assessment of the targets. The training of the models is performed from the leaves to the root, stacking the models' predictions as new inputs.


\subsection{Meta-learning for Multi-target regression}

During the literature research, we did not find any papers employing \acrshort{mtl} for \acrshort{mtr}. However, in some studies, the authors investigated similar problems, such as \acrfull{mlc} problems.

Considering $\mathcal{L}$ the set of labels, differently from Single-label classification task, which there is just one label $L_i \in \mathcal{L}$ to predict for each dataset's example,  in \acrshort{mlc} tasks the examples are associated with more than one label, i.e., it is necessary to learn how to associate the example with a subset of $\mathcal{L}$.

Similarly to the problem investigated in this paper, many \acrshort{mlc} methods~\cite{tsoumakas2007multi} were proposed, but there is few research concerning when each method is more efficient. 

To select a \acrshort{mlc} method and configure their hyperparameters for a given dataset, de Sá \textit{et al.}~\cite{de2017towards} applied Evolutionary Algorithms (EA). This study was carried using $31$ MLC methods, in $3$ different datasets. The EA selection outperformed or at least draw the baselines in most of the cases. 
Also in this direction, the pioneering research based on~\acrshort{mtl} was done by Chekina \textit{et al.}~\cite{chekina2011meta}. They evaluated $11$ different multi-label methods, grouping them into: Single-Classifier Algorithms and Ensemble-Classifier Algorithms. They performed experiments in $12$ datasets of \acrshort{mlc} from the literature. The results showed that employing \acrshort{mtl} to select one method in \acrshort{mlc} tasks 
is promising, since in most of the  experimented cases, to apply the recommendation through \acrshort{mtl} was better than selecting one method for all tasks or selecting it randomly.

\acrshort{mlc} tasks are similar to \acrshort{mtr} tasks, since both deal with the prediction of multiple targets using a common set of features. The main difference is the type of the predicted variable: while in \acrshort{mlc} the targets are binary, in \acrshort{mtr} the outputs are continuous. Indeed, both tasks can be seen as a more general learning task of multi-target prediction with different types of variables to predict~\cite{Spyromitros2016}. Therefore, given that \acrshort{mtl} was successfully applied to select \acrshort{mlc} methods, it is significant to experiment \acrshort{mtl} to select \acrshort{mtr} methods.


\section{Material and Methods}
\label{sec:materials}

\begin{figure}[ht!]
    \centering
    \includegraphics[scale=0.48]{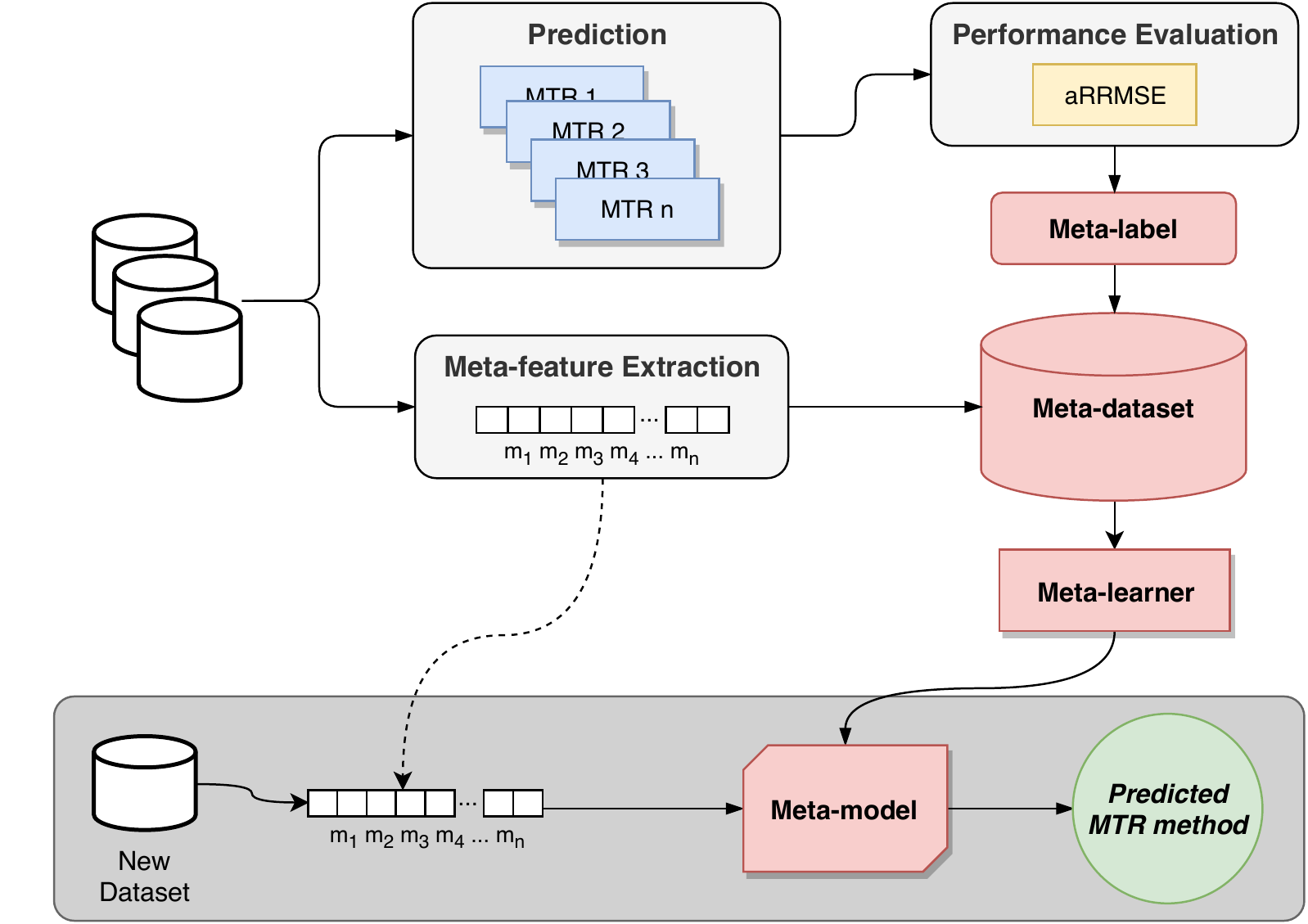}
    \caption{Overview of the procedure to select a Multi-target Regression method through Meta-learning.}
    \label{fig:approach}
\end{figure}

Fig~\ref{fig:approach} provides an overview of the adopted experimental methodology. First, we performed exhaustive experiments evaluating all the \acrshort{mtr} methods in all available datasets. We also identified the best method for each dataset, selecting the one with the smallest \acrfull{rrmse}. This information is used to define the meta-label.
At the same time, a set of measures, named meta-features, are also extracted to describe each {dataset}. 
We then unify the meta-feature values with the meta-labels to compose our meta-dataset.
Then, we can employ ML algorithms to predict the best \acrshort{mtr} method for a new unseen dataset. 
The next subsections describe each one of these processes with details.

\subsection{Datasets}

In the experiments, the meta-dataset was composed of $648$ benchmarking synthetic datasets\footnote{The generated datasets are available for download in: http://www.uel.br/grupo-pesquisa/remid/?page\_id=145}
, generated by following the procedure described in~\cite{mastelini2018benchmarking}. 
We used synthetic datasets to overcome the lack of real datasets that meet specifics scenarios of inter-targets dependencies, complexity levels from the input to output relations, and cover a different number of input features and targets. 
To create a wide possibility of datasets, the parameters of the dataset generator assumed the values presented in Table~\ref{tab:generator}. The numeric targets were built upon math expressions of identity, quadratic, and cubic functions, or their combination. \\


\begin{table}[h!]
    \centering
    \caption{Parameters used to generate synthetic base level datasets.}
    \label{tab:generator}
    \begin{tabular}{clc}
        \toprule
        \textbf{Symbol} & \textbf{Hyperparameter} & \textbf{Values} \\
        \midrule
        N & Number of instances & \{500,\:1000\} \\
        m & Number of features & \{15,\:30,\:45,\:60,\:75,\:90\} \\
        d & Number of targets & \{3,\:6\} \\
        g & Generating groups & \{1,\:2\} \\
        $\eta$ & \% Instances affected by noise & \{1,\:5,\:10\} \\
        \bottomrule
    \end{tabular} 
\end{table}


\subsection{Meta-features}

Each base-level dataset is represented by a vector of characteristics, the meta-features. In~\cite{Brazdil:2009} the authors list some requirements that a meta-features must follow: they need to have good discriminative power, their extraction should not be computational complex and the number of meta-features should not be large to avoid overfitting.

In our meta-level experiments, a set of $58$ meta-features were explored. They included measures from different categories: statistical information about the dataset (STAT), correlation between attributes and targets (COR), performance metrics related to a linear regression (LIN), distribution of the dataset (DIM) and smoothness of the data (SMO)~\cite{lorena2018data,Mantovani:2019}. 

It is important to mention that some of these meta-features were designed for problems with one single 
output. Since we are dealing with multi-target problems, the real value of the meta-features were aggregated, given that a meta-feature is extracted for each target. To overcome this problem, the meta-feature was extracted for each target, then the average, standard deviation, maximum and minimum was added to the set of meta-features~\cite{Rivolli:2018}. Most of the meta-features values were extracted using the R package \texttt{ECoL}~\cite{lorena2018data}.
A complete list of the meta-features used in the experiments is presented in Table \ref{tab:metafeatures}.


\begin{table}[h]
\centering
\caption{Type, acronym, aggregation function (when applied) and description of meta-features used in the experiments.}
\label{tab:metafeatures}
\resizebox{0.48\textwidth}{!}{%
\begin{tabular}{cccl}
\toprule
\multirow{2}{*}{\textbf{Type}} & \multirow{2}{*}{\textbf{Acronym}} & \textbf{Aggregation} & \multicolumn{1}{c}{\multirow{2}{*}{\textbf{Description}}} \\
& & \textbf{Functions} & \\
\midrule
\multirow{6}{*}{STAT} & n.samples & - & Number of samples \\
& n.attributes & - & Number of attributes \\
& n.targets & - & Number of targets \\
& target.ratio & - & Ratio between targets and attributes \\
& pc{[}1-3{]} & - & \begin{tabular}[c]{@{}l@{}}First three components of \\ \:\:the Principal Components Analysis\end{tabular} \\

\midrule

\multirow{5}{*}{DIM} & T2& - & Average number of samples per dimension \\
& T3 & - & \begin{tabular}[c]{@{}l@{}}Average intrinsic dimensionality\\  \:\:per number of examples\end{tabular} \\
& T4 & - & Intrinsic dimensionality proportion \\

\midrule
\multirow{5}{*}{COR} & 
cor.targets & \{avg,max,min,sd\} & Correlation between targets \\
& C1  & \{avg,max,min,sd\} & Maximum feature correlation to the output \\
& C2  & \{avg,max,min,sd\} & Average feature correlation to the output \\
& C3 & \{avg,max,min,sd\} & Individual feature efficiency \\
& C4 & \{avg,max,min,sd\} & Collective feature efficiency \\
\midrule
\multirow{4}{*}{LIN} & 
regr.L1 & \{avg,max,min,sd\} & \begin{tabular}[c]{@{}l@{}}Distance of erroneous instances\\  \:\:to a linear classifier\end{tabular} \\
& regr.L2  & \{avg,max,min,sd\} & Training error of a linear classifier \\
& regr.L3 & \{avg,max,min,sd\} & Nonlinearity of a linear classifier \\
\midrule
\multirow{4}{*}{SMO} & 
S1  &\{avg,max,min,sd\} & Smoothness of the output distribution \\
& S2  & \{avg,max,min,sd\} & Smoothness of the input distribution \\
& S3 & \{avg,max,min,sd\} & Error of a k-nearest neighbor regressor \\
& S4 & \{avg,max,min,sd\} & Non-linearity of nearest neighbor regressor \\ 
\bottomrule
\end{tabular}%
}
\end{table}


\subsection{Meta-labels}

\acrshort{st} approach and three \acrshort{mtr} methods were explored in experiment: \acrshort{sst}, \acrshort{erc}~\cite{Spyromitros2016} and \acrshort{motc}~\cite{Mastelini2018}.
Even being the most simple, the \acrshort{st} approach was included in the experimental setup because it can perform better than~\acrshort{mtr} methods in problems with limited inter-target dependency. On the other hand, the other three~\acrshort{mtr} methods were selected because they offer a proper trade-off between performance and time complexity, as concluded from~\cite{mastelini2018benchmarking}.

These four different methods mentioned above were executed 
for every single base-level dataset. Their induced models were assessed in terms of~\acrfull{rrmse} evaluation measure defined in Equation \ref{eq:arrmse}, where $N$ represents the number of instances, and $y$, $\hat{y}$ and $\overline{y}$ represent, respectively, the true, predicted and mean values of the target.

\acrlong{svm} was used as base regressor, performing a k-Fold~\acrfull{cv} resampling strategy, with $k = 10$. \acrshort{svm} was chosen as base regressor due to its usage in the most of MTR Problem transformation literature \cite{Tsoumakas2014,Mastelini2017,Santana2017,sanatana2018,Mastelini2018}. The method with the smallest \acrshort{rrmse}~\cite{Borchani2015} was chosen as the best multi-target method for every dataset. The experiments were performed using the \texttt{mtr-toolkit}\footnote{https://github.com/smastelini/mtr-toolkit},
implemented in R. Thus, our meta-dataset was a multi-class meta-label with four different levels indicating the best \acrshort{mtr} method or ST regression. The class distribution (\%) in the meta-dataset is also presented in Table \ref{tab:specification}. 

\begin{equation}
    \text{aRRMSE} = \frac{1}{d}\sum^{d}_{t=1} \sqrt{\frac{\sum_{i=1}^{N} (y_{t}^{i} - \hat{y}_{t}^{i})^2} {\sum_{i=1}^{N} (y_{t}^{i} - \overline{y})^2}}
    \label{eq:arrmse}
\end{equation}


\begin{table}[h]
\centering
\caption{Specification of the meta-dataset used in experiments}
\label{tab:specification}
\begin{tabular}{@{}crrrrc@{}}
\toprule
& ERC & MOTC & SST & ST & \textbf{Total} \\ 
\midrule
examples & 166 & 89 & 362 & 31 & 648 \\
\% & 25.6 & 13.7 & 55.8 & 4.9 & 100 \\ 
\bottomrule
\end{tabular}
\end{table}


\subsection{Meta-learners}

Four \acrshort{ml} algorithms, with different learning biases, were used as meta-learners: \acrfull{nb}~\cite{russell2016artificial}, \acrfull{rf}~\cite{Breiman:2001}, \acrfull{svm}~\cite{Vapnik:1995} and \acrfull{xgb}~\cite{Chen2016}. These algorithms were selected due to their widespread use and capacity of high-performance models induction. The k-Fold \acrshort{cv} resampling methodology was also adopted in the meta-level of the experiments to assess the predictive performance of the meta-learners, with $k = 10$ folds. 
All the \acrshort{ml} algorithms were implemented in R, using the \texttt{mlr} package and their correspondent default hyperparameters.


\subsection{Evaluation measures and baselines}

Seven evaluation metrics were used to assess the predictive performance of the induced models:  Accuracy,  Balanced per class accuracy, Precision, Recall, F-score (f1), Sensitivity and Specificity.

Besides, we used two different baselines from the \acrshort{mtl} literature for comparisons: a model that always recommends the majority class for the whole dataset (\texttt{Majority}) and a model that provides random recommendations (\texttt{Random}). These baselines are widely used to endorse the need for a recommendation system~\cite{Brazdil:2009}. 
Also, we used an upper-bound as the ground-truth (\texttt{Truth}).



\section{Results and Discussion}
\label{sec:results}

The results were organized starting by exposing the results regarding the predictive performance of meta-models from different ML algorithms. Afterward, based on the RF meta-model performance, the meta-features were compared and discussed. Finally, some contributions and open issues related to \acrshort{mtl} and \acrshort{mtr} were presented.

\subsection{Predictive Performance}

\begin{figure}[h!]
    \centering
    \includegraphics[width=.42\textwidth]{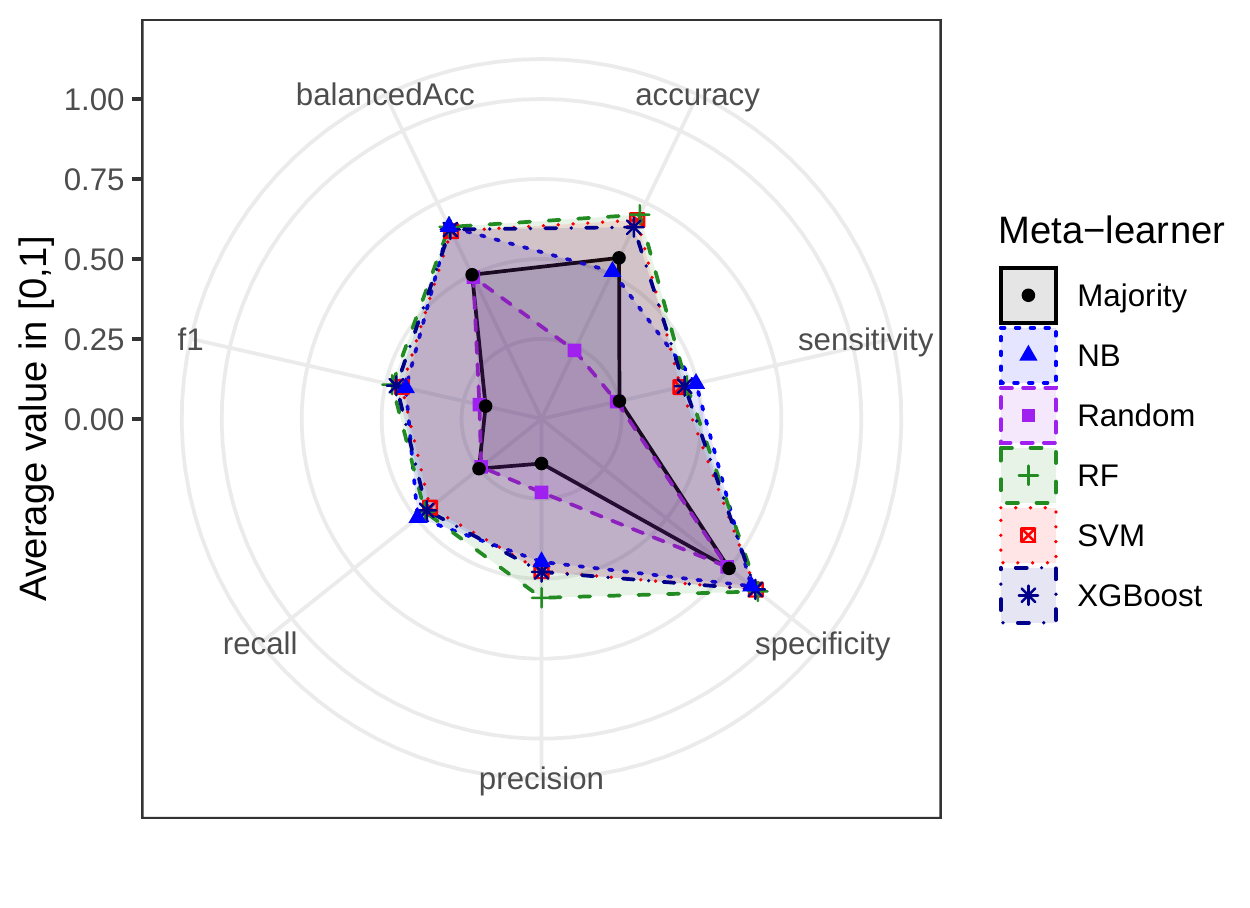}
    \caption{Performance of the meta-models}
    \label{fig:radarchart}
\end{figure}

The predictive performance obtained by the four meta-learners and the baselines are presented as a radar chart in Fig.~\ref{fig:radarchart}. 
In this figure, each line represents a meta-model and each vertex its related to a different performance measure. The larger the area in the radar chart, the better the meta-model.


Looking at the radar chart, it is possible to observe that all meta-models had a superior performance than \texttt{Random} baseline for all metrics. The same occurs for \texttt{Majority}, except for accuracy with \acrshort{nb}, since \texttt{Majority} has $55.8$\% of accuracy, whereas the \acrshort{nb} meta-model achieved $51.08$\%.
Still for this metric, \acrshort{rf} obtained the best results with $70.83$\% of accuracy. Following the \acrshort{rf}, the \acrshort{svm} achieved $68.9$\% and \acrshort{xgb} was the third, with $66.51$\% of accuracy. The only metric that \acrshort{rf} meta-model did not obtain the higher value was Sensitivity, when \acrshort{nb} was the best with $0.49$. Regarding the other evaluation metrics, \acrshort{rf} achieved the best results, with $0.86$ of Specificity, $0.55$ of Precision, $0.46$ of Recall, $0.47$ of F1 and $66.59$\% of Balanced per class accuracy.


Although three of four meta-models overcame the baselines for all metrics, the predictive performance did not achieve high values, which might be related to the meta-dataset imbalance problem. 
However, the superiority of the \acrshort{mtl} recommending system regarding the baselines was confirmed by statistical tests. 
We used the Friedman test, with a significance level of $\alpha = 0.05$. The null hypothesis is that the recommendation by the meta-models and by the baselines are similar.
Anytime the null hypothesis is rejected, the Nemenyi post hoc test can be applied, stating that the performance of the two approaches are significantly different if their corresponding average ranks differ by at least a \acrfull{cd} value. When multiple algorithms are compared in this way, a graphic representation can be used to represent the results with the \acrshort{cd} diagram, as proposed by Dem{\v{s}}ar~\cite{demvsar2006statistical}. 

\begin{figure}[h] \centering \begin{tikzpicture}[xscale=1.2]
\node (Label) at (1.0087716157056061, 0.7){\tiny{CD=0.35}}; 
\draw[decorate,decoration={snake,amplitude=.4mm,segment length=1.5mm,post length=0mm},very thick, color = black] (0.8571428571428571,0.5) -- (1.160400374268355,0.5);
\foreach \x in {0.8571428571428571, 1.160400374268355} \draw[thick,color = black] (\x, 0.4) -- (\x, 0.6);
 
\draw[gray, thick](0.8571428571428571,0) -- (6.0,0); 
\foreach \x in {0.8571428571428571,1.7142857142857142,2.5714285714285716,3.4285714285714284,4.285714285714286,5.142857142857143,6.0} \draw (\x cm,1.5pt) -- (\x cm, -1.5pt);
\node (Label) at (0.8571428571428571,0.2){\tiny{0}};
\node (Label) at (1.7142857142857142,0.2){\tiny{1}};
\node (Label) at (2.5714285714285716,0.2){\tiny{2}};
\node (Label) at (3.4285714285714284,0.2){\tiny{3}};
\node (Label) at (4.285714285714286,0.2){\tiny{4}};
\node (Label) at (5.142857142857143,0.2){\tiny{5}};
\node (Label) at (6.0,0.2){\tiny{6}};
\draw[decorate,decoration={snake,amplitude=.4mm,segment length=1.5mm,post length=0mm},very thick, color = black](3.0829365079365085,-0.25) -- (3.3066137566137566,-0.25);
\draw[decorate,decoration={snake,amplitude=.4mm,segment length=1.5mm,post length=0mm},very thick, color = black](3.6623015873015867,-0.4) -- (3.766931216931217,-0.4);
\node (Point) at (2.3015873015873014, 0){};\node (Label) at (0.5,-0.65){\scriptsize{\texttt{Truth}}}; \draw (Point) |- (Label);
\node (Point) at (3.1329365079365084, 0){};\node (Label) at (0.5,-0.95){\scriptsize{RF}}; \draw (Point) |- (Label);
\node (Point) at (3.17989417989418, 0){};\node (Label) at (0.5,-1.25){\scriptsize{SVM}}; \draw (Point) |- (Label);
\node (Point) at (4.699735449735449, 0){};\node (Label) at (6.5,-0.65){\scriptsize{\texttt{Random}}}; \draw (Point) |- (Label);
\node (Point) at (3.716931216931217, 0){};\node (Label) at (6.5,-0.95){\scriptsize{\texttt{Majority}}}; \draw (Point) |- (Label);
\node (Point) at (3.7123015873015865, 0){};\node (Label) at (6.5,-1.25){\scriptsize{NB}}; \draw (Point) |- (Label);
\node (Point) at (3.2566137566137567, 0){};\node (Label) at (6.5,-1.5499999999999998){\scriptsize{XGBoost}}; \draw (Point) |- (Label);
\end{tikzpicture}
\caption{Comparison of the aRRMSE values obtained by meta-models when recommending MTR methods according to the Nemenyi test.  Groups that are not significantly different ($\alpha$= 0.05 and CD = 0.35) are connected.}
\label{fig:nemenyi}
\end{figure}
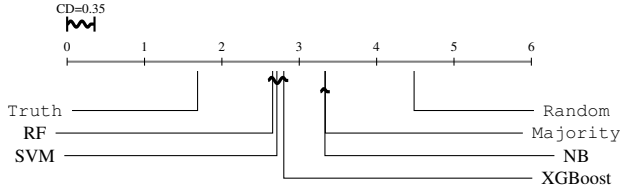

The meta-models (\acrshort{rf}, \acrshort{svm}, \acrshort{xgb}, \acrshort{nb}) were compared with \texttt{Truth} (expected method), the \texttt{Majority} (which always predicts the \acrshort{sst}) and \texttt{Random} (the random selection of a method for each dataset), using the \acrshort{rrmse} of the prediction as performance metric.
This analysis is shown in Fig. \ref{fig:nemenyi}, using the results from the Nemenyi test.

As exposed in Fig. \ref{fig:nemenyi}, no solution was similar to the \texttt{Truth}, which was expected due to the predictive performance. However, the \acrshort{rf}, \acrshort{svm}, \acrshort{xgb} are connected, which means they were similar and superior the baselines \texttt{Majority} and \texttt{Random}. This fact supports the benefit of using \acrshort{mtl} recommending system in comparison to select a specific algorithm for every dataset or select it randomly.

\subsection{Relative importance of the meta-features }

\begin{figure*}[ht!]
    \centering
    \includegraphics[width=\textwidth]{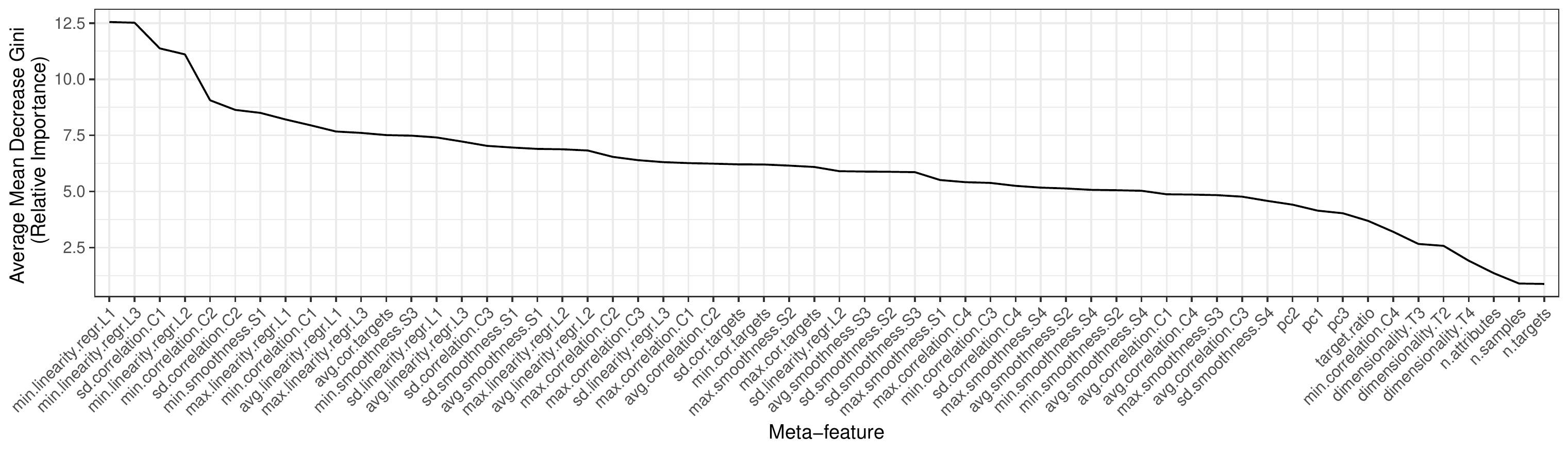}
    \caption{Average relative importance of the meta-features obtained from RF importance.   The names of the meta-features in the x-axis follow the acronyms presented in Table \ref{tab:metafeatures}.}
    \label{fig:rf_importance}
\end{figure*}

\acrlong{rf} meta-model was used to assess the importance of each meta-feature by using the RF Feature Importance metric. This metric is calculated by permuting the values of a feature in the \acrfull{oob} samples and recalculating the \acrshort{oob} error in the whole ensemble. In other words, if substituting the values of a meta-feature by random values results in error increase, this meta-feature is considered important. Otherwise, if the error decreases, the resulting importance is negative. Thus, the meta-feature is considered not important and should be removed from modeling. This procedure could be performed for each meta-feature toward explaining its impact~\cite{Breiman:2001}. Fig. \ref{fig:rf_importance} shows the meta-feature importance for the meta-dataset.  


\acrlong{cor} and \acrlong{lin} meta-features achieved the higher values of importance, especially the Minimum value of distance of erroneous instances to a linear classifier ($12.54$), Minimum value of non-linearity of a linear classifier ($12.51$) and the Standard Deviation of the Maximum feature correlation to the output ($11.37$). Once the \acrshort{mtr} method tries to explore the correlation between the features and the targets in different ways, their selection makes sense.
The number of targets, attributes and samples had low importance. This might have occurred because these meta-features did not influence in the predictive performance, showing that the \acrshort{mtr} methods used in the experiments can deal with different numbers of targets, attributes and samples in the same way.

\subsection{Insights and open issues}

 It is important to highlight the meta-label attribution was straightforward related to the highest predictive performance (low \acrshort{rrmse}) based on the ranking of methods. Differences between the predictive performance of the \acrshort{mtr} methods, independent of their magnitude, were not considered while building the meta-dataset. 
 
 Alternatively, the meta-label assessment could be performed by indicating two or more methods suitable to solve a given problem in the case of no statistical difference between their performances. However, this scenario poses an additional challenge to deal with a multi-label problem in the meta-level of the recommending system.
 
 Another important issue was the fact of meta-label assessment was made regarding only low predictive error of \acrshort{mtr} methods. In some cases, e.g., Online Multi-target Regression \cite{osojnik2018tree}, the most proper method concerns to address a trade-off among predictive performance, memory, and time cost when predicting the output. This scenario demands additional information, as well as complexity, toward identifying the best \acrshort{mtr} method to be learned by the recommending system.




\section{Conclusions and Future Work}
\label{sec:conclusion}

In this study, a framework for recommending \acrshort{mtr} methods using meta-learning was presented. A meta-dataset, composed with $648$ datasets used for \acrshort{mtr} methods benchmark, was created for the induction of meta-models toward predicting the best one for a given dataset. Experiments performed with the meta-dataset and four meta-learners led to 70.83\% of accuracy with RF, the best recommender. Besides, it overcame the baselines, and statistical tests showed that the recommendation system was better than select one for every task or selecting a method randomly. The analysis of meta-feature importance revealed that correlation between targets and error of a linear classifier were the most useful features to predict the performance of a \acrshort{mtr} method for a given unseen dataset.

As future work, besides implementing more meta-features, we intend to use more \acrshort{mtr} benchmarking datasets, in order to improve the generalization capability of the meta-models. Also, we expect to apply \acrshort{mlc} to predict the \acrshort{mtr} method and its base regressor. Further information related to the memory and time cost will be used to match the requirement of different scenarios, e.g., Online \acrshort{mtr}.


\section*{Acknowledgements}
The authors would like to thank the financial support of Coordination for the Improvement of Higher Education Personnel (CAPES) - Finance Code 001 -, the National Council for Scientific and Technological Development (CNPq) of Brazil - Grant of Project 420562/2018-4 - and S{\~a}o Paulo Research Foundation (FAPESP) - grant \#2018/07319-6.

\bibliographystyle{./bibliography/IEEEtran}
\bibliography{./bibliography/ref}

\end{document}